\documentclass[10pt, a4paper]{article}

\usepackage[final]{lrec-coling2024} 

\usepackage{amsfonts}
\usepackage{algorithm}
\usepackage{algpseudocode}
\usepackage{multirow}
\usepackage{multicol}
\usepackage{booktabs}
\usepackage{tikz}
\usepackage{pgfplots}

\title{MACT: Model-Agnostic Cross-Lingual Training for Discourse Representation Structure Parsing}

\name{Jiangming Liu} 

\address{Yunnan University \\
Yunnan Key Laboratory of Intelligent Systems and Computing, Yunnan University \\
         jiangmingliu@ynu.edu.cn\\}

\abstract{
Discourse Representation Structure (DRS) is an innovative semantic representation designed to capture the meaning of texts with arbitrary lengths across languages.
The semantic representation parsing is essential for achieving natural language understanding through logical forms.
Nevertheless, the performance of DRS parsing models remains constrained when trained exclusively on monolingual data.
To tackle this issue, we introduce a cross-lingual training strategy.
The proposed method is model-agnostic yet highly effective. It leverages cross-lingual training data and fully exploits the alignments between languages encoded in pre-trained language models. 
The experiments conducted on the standard benchmarks demonstrate that models trained using the cross-lingual training method exhibit significant improvements in DRS clause and graph parsing in English, German, Italian and Dutch. Comparing our final models to previous works, we achieve state-of-the-art results in the standard benchmarks. Furthermore, the detailed analysis provides deep insights into the performance of the parsers, offering inspiration for future research in DRS parsing. \textit{\textbf{We keep updating new results on benchmarks to the appendix.}}
 \\ \newline  }

\begin{document}

\maketitleabstract

\section{Introduction}
Discourse Representation Structure (DRS) is a novel semantic representation rooted the Discourse Representation Theory (DRT; \citealt{kamp1993discourse}), which has been developed to encompass a wide range of linguistic phenomena, such as discourse relations, the interpretation of pronouns, and temporal expressions, in texts spanning arbitrary lengths and multiple languages. 
Researchers have proposed various models for parsing texts into DRS representations in the form of boxes \cite{bos-2008-wide,bos-2015-open,evang-2019-transition}, clauses \cite{van-noord-etal-2018-exploring,van-noord-etal-2019-linguistic,liu-etal-2019-discourse-representation,van-noord-etal-2020-character,wang-etal-2021-input,liu-etal-2021-universal}, trees \cite{liu-etal-2018-discourse,liu-etal-2019-discourse,liu-etal-2021-universal} and graphs \cite{poelman-etal-2022-transparent,wang-etal-2023-pre}.

The widespread success of pre-trained language models (PLMs) in various NLP tasks has ushered in a new training paradigm where semantic models are built on PLMs \cite{bai-etal-2022-graph,sun2023sql}.
Likewise, recent DRS parsing models are trained using PLMs \cite{van-noord-etal-2020-character,wang-etal-2021-input,wang-etal-2023-pre}. Building an universal model that can parse multiple language text is necessary \cite{vilares-etal-2016-one,de-lhoneux-etal-2018-parameter,kondratyuk-straka-2019-75}.

\begin{figure}[!tp]
    \centering
    \includegraphics[width=200pt]{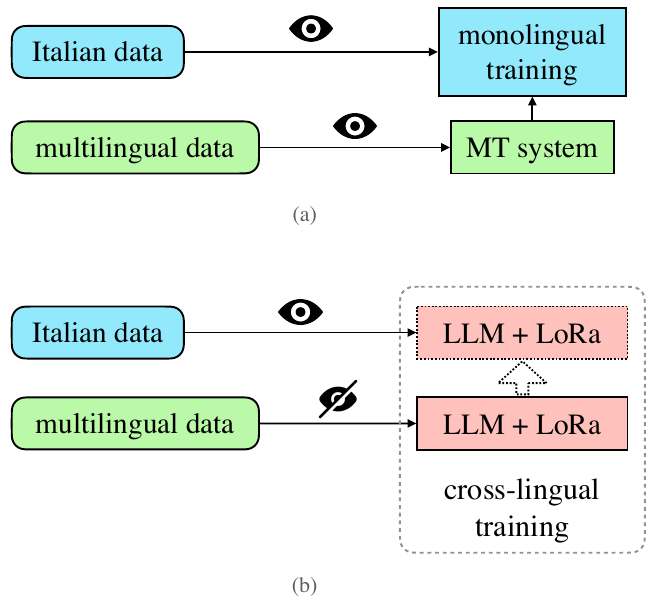}
    \caption{
    Training of an Italian semantic parser. (a) Monolingual training with multilingual data using machine translation systems. (b) cross-lingual training without language identifications.
    }
    \label{front}
\end{figure}

However, recent language-specific DRS parsing models are trained exclusively on monolingual data, overlooking the valuable insights available from the interaction across languages.
As shown in the blue part of Figure \ref{front}(a), Italian DRS parsing models are typically trained exclusively on Italian data, without considering data from other languages. 
Despite the potential use of machine translation systems to include more training data from other languages by translating them into Italian, as shown in Figure \ref{front}(a), the training is still monolingual, and the well-trained models lack consistency and are strongly influenced by the quality of translations. Due to the application of machine translation system, we have to know which language that we use to enhance the models.

A variety of models have been proposed to enhance pre-trained language models (PLMs) with cross-lingual generalization capabilities \cite{lample2019cross,conneau-etal-2020-unsupervised,liu-etal-2020-multilingual-denoising,xue-etal-2021-mt5,tang-etal-2021-multilingual,shliazhko2022mgpt,lin-etal-2022-shot,muennighoff-etal-2023-crosslingual}.
\citet{muennighoff-etal-2023-crosslingual} gather diverse multilingual supervised datasets to fine-tune PLMs within a multitask learning framework. Leveraging PLMs, the fine-tuned models effectively capture rich contextual representations, leading to state-of-the-art performance in these tasks. Simultaneously, PLMs are enriched with additional task-specific information, such as cross-lingual generalization.

Motivated by the cross-lingual generalization in PLMs, we introduce cross-lingual training for discourse representation structure parsing. 
Our proposed method, which is model-agnostic and straightforward yet effective, involves the collection of multilingual training data to build language-specific semantic parsers. 
As shown in Figure \ref{front}(b), the cross-lingual training requires language-specific training data (Italian data) and a set of multilingual training data.
Instead of translating the multiple languages via the MT systems, the cross-lingual training approach directly utilizes the multilingual data to train the models. Thanks to the cross-lingual generalization of PLMs, we do not have to identify the languages used in the training instances within the multilingual training data.

The experiments are conducted on the Parallel Meaning Bank (PMB; \citealt{abzianidze-etal-2017-parallel}), a standard benchmark for DRS parsing.
We employ the cross-lingual training to build models for both DRS clause parsing and DRS graph parsing.
The experiments show that the final models, trained with our proposed cross-lingual training method, achieve state-of-the-art results in DRS parsing for English, German, Italian, and Dutch.

In addition, we make a detailed analysis of the outputs generated by our final models, highlighting the significance of controlling the DRS parser to ensure the the outputted DRSs are well-formed.
All the models in the experiments do not explicitly impose constraints on well-formed parsing, yeTt our final models consistently generate more well-formed DRSs. The contributions are summarized as following:
\begin{itemize}
    \item We introduce cross-lingual training for DRS parsing, without the need for bitexts provided by machine translation systems or specifying the languages of the training instances.
    \item The proposed cross-lingual training approach is simple yet effective and can be applied to general models.
    \item The final models, enhanced with our proposed cross-lingual training, achieve state-of-the-art results in both DRS clause parsing and DRS graph parsing.
    \item We make a detailed analysis of DRS parsing, revealing that DRS parsing is susceptible to the issue of semantic over-generation, especially in the case of short texts.
\end{itemize}
Our dataset and code are available at \url{https://github.com/LeonCrashCode/DRS-Cross-Lingual-Training}.

\section{Related Work}
\subsection{Discourse Representation Structure Parsing}

In recent years, there has been a growing interest in the development of DRS parsing models. Early seminal work \cite{bos-2008-wide} introduced an open-domain semantic parser that generates DRS representations in box form by leveraging the syntactic analysis offered by a robust CCG parser \cite{curran-clark-bos:2007:PosterDemo}.

The first data-driven DRS parser was proposed by
\citet{le-zuidema:2012:PAPERS} based on a graph representation. The availability of annotated corpora has subsequently enabled the exploration of neural models. 
\citet{liu-etal-2018-discourse,liu-etal-2019-discourse} conceptualize DRS parsing as a tree structure prediction problem which they model with a series of encoder-decoder architectures. 
\citet{van-noord-etal-2018-exploring,van-noord-etal-2019-linguistic,van-noord-etal-2020-character} adapt sequence-to-sequence models with LSTM units and transformers to parse DRSs in clause form. 
\citet{poelman-etal-2022-transparent} convert DRSs into graph forms similar to AMR graphs based on universal dependencies, eliminating variable bindings. Based on DRS graphs, \citet{wang-etal-2023-pre} employ language modeling techniques to build a pre-trained model concentrating on DRSs. Our cross-lingual training approach is general and can be adapted to DRS in all these forms.


\begin{figure*}[!tp]
    \centering
    \includegraphics[width=440pt]{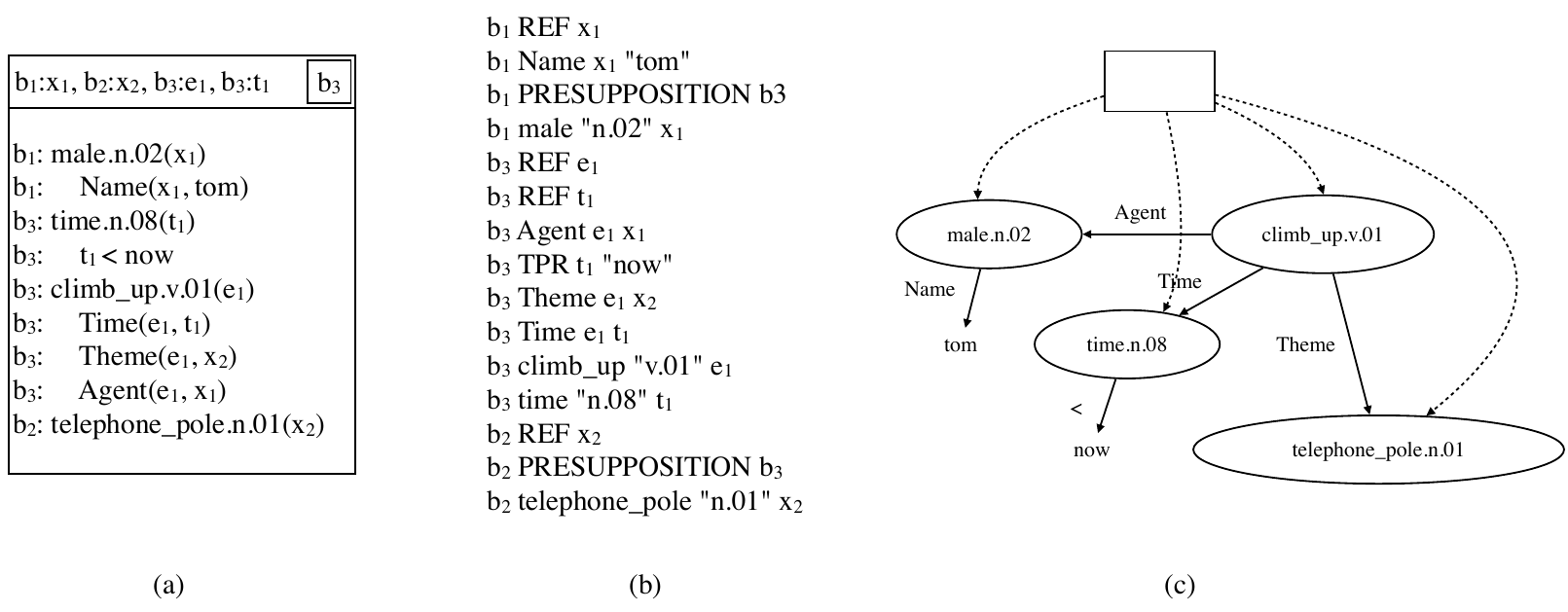}
    \caption{
    DRSs of the English sentence ``Tom climbed up the telephone pole'' in (a) box form, (b) clause form, and (c) graph form.
    }
    \label{drs}
\end{figure*}
\subsection{Cross-Lingual Semantic Parsing}

The idea of using English annotations to address resource scarcity in other languages through translational equivalences is not a new one.
Various methods have been proposed in the literature within the general framework of annotation projection. \cite{yarowsky-ngai-2001-inducing,hwa2005bootstrapping,pado2005cross,pado2009cross,akbik2015generating,evang2016cross,damonte2018cross,zhang2018cross} which primarily involves projecting existing annotations from the source-language text to the target language.
There is another line of researches concentrating on translation systems that translate multiple languages into the desired language, which is then used to train the models \cite{conneau-etal-2018-xnli,yang-etal-2019-paws,huang-etal-2019-unicoder,sherborne-lapata-2022-zero}.

Our cross-lingual training follows the model transfer approach commonly adopted in the literature \cite{cohen2011unsupervised,mcdonald2011multi,sogaard2011data,wang-manning-2014-cross,sherborne-lapata-2022-zero,zhang-etal-2023-xsemplr,wang-etal-2023-pre}, where model parameters are shared across languages.


\subsection{Parameter-Efficient Fine-Tuning}

Fine-tuning pre-trained language models is a prevalent paradigm in recent natural language processing tasks. Nevertheless, fine-tuning all model parameters is resource-intensive and time-consuming. To address these challenges, researchers have introduced several parameter-efficient fine-tuning methods. Within the general framework of prompt tuning \cite{li-liang-2021-prefix,DBLP:journals/corr/abs-2103-10385,lester-etal-2021-power}, a multitude of techniques have been proposed in the literature. These approaches aim to design or search for suitable prompts or prompt embeddings to tailor pre-trained models for specific downstream tasks.

Our works align with the framework commonly adopted in the literature, which incorporates lightweight and trainable structures into pre-trained models \cite{houlsby2019parameter} or adds the compressed parameters tailored for downstream tasks \cite{hu2021lora,ansell-etal-2022-composable,xu-etal-2021-raise}. Moreover, \citet{he2021towards} introduce a unified framework that establishes connections between these approaches.

\begin{figure*}[!tp]
    \centering
    \includegraphics[width=440pt]{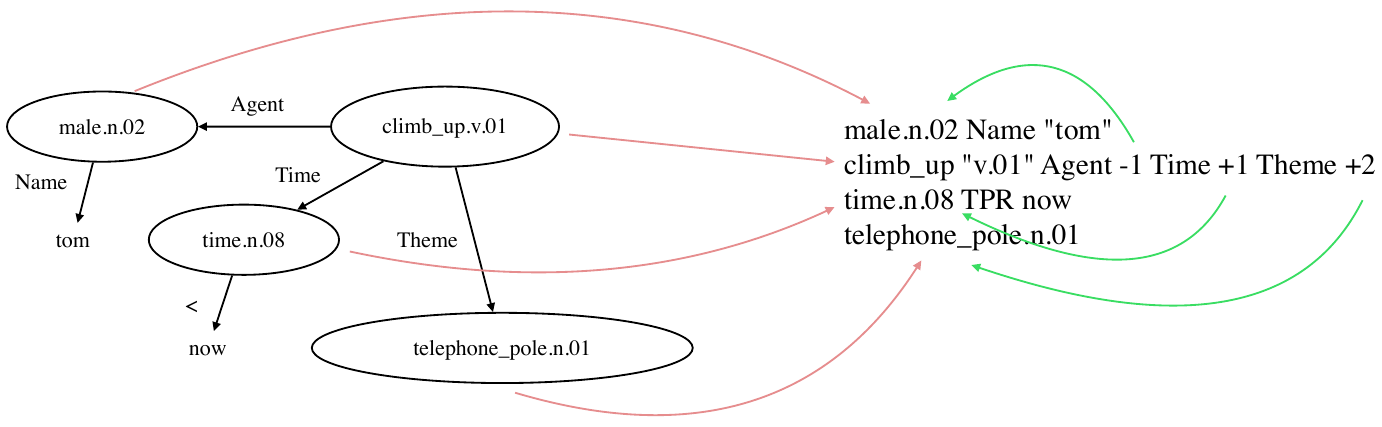}
    \caption{Examples of DRSs in sequential graph form. The red arrow lines indicate the mapping from nodes to corresponding items, and the green arrow lines indicate the argument positions.}
    \label{seq}
\end{figure*}

\section{Discourse Representation Structure}

Discourse Representation Structures (DRSs) serve as the fundamental meaning-carrying units in Discourse Representation Theory (DRT). 
They are structured as nested boxes, recursively representing semantics within and across sentences, as illustrated in Figure \ref{drs}(a). 
These boxes consist of two layers: the upper layer containing variables and a box label, and the lower layer containing semantic conditions. 
However, while box-style DRSs are easy to read, they are not particularly amenable to modeling. 
As a result, box-style DRSs are often transformed into alternative structures.

\subsection{Clauses}
DRS can be converted into a set of clauses, with each clause serving as a fundamental semantic unit. The transformation between the box notation and the clauses is straightforward, as conditions, relations, and variables are placed in the clause and preceded by the label of the box they originate from. As shown in Figure \ref{drs}(b), each clause starts with a box label, followed by a relation and the corresponding variables. 
These clauses can represent unary or binary relations. 
The transformation from boxes to clauses removes the nested structure information present in the boxes.\footnote{We adopt the transformation approach used in previous work \cite{van-noord-etal-2018-exploring}.}

\subsection{Graph}
DRS can be converted into a graph. In the graphs, nodes represent predicates, entities, or dummy nodes indicating boxes, while the edges have three distinct types: semantic relations between predicates (entities), discourse relations between dummy nodes, or predicate positions indicating the boxes (dummy nodes) where the predicates occur.

As shown in Figure \ref{drs}(c), the predicate \texttt{climb\_up.v.01} is connected to the predicate \texttt{male.n.02} with the \texttt{Agent} edge, and the predicate \texttt{male.n.02} is connected to the entity \texttt{tom}, showing that the agent of the event \texttt{climb\_up.v.01} is the \texttt{male.n.02} called \texttt{tom}. All the predicates are connected to a dummy box node with dashed line, showing that they are placed in the same box (predicate position). Therefore, the variable bindings are removed by the box-to-graph transformation.

However, the transformation eliminates the presupposition resolver, that is used for the truth-condition logic inference, by removing the label of the box used to indicate the interpretation positions of the predicates.
In Figure \ref{drs}(a), the predicate \texttt{male.n.08($x_1$)} and the predicate \texttt{telephone\_pole.n.02($x_2$)} are interpreted in the box $b_1$ and the box $b_3$, respectively, but the corresponding nodes in Figure \ref{drs}(c) are connected to the same box dummy node. 

\section{Methods}
In this section, we frame DRS parsing as conditional generation and introduce the parsing model with parameter-efficient fine-tuning. Then, we present the model-agnostic cross-lingual training method.

\subsection{DRS parsing}
We model DRS parsing in clause and graph forms as a conditional generation task. The input is a sequence of words, $X = [x_0, x_1, ..., x_m]$, where $m$ is the length of the input text. Given the input sequence, the models conditionally generate a sequence of DRS symbols, $Y = [y_0, y_1, ..., y_l]$, where $l$ is the length of the output.

\paragraph{Sequential Clause Form.}
The transformation of DRSs in clause form into a sequence of symbols is straightforward. Starting from the top and moving downwards, we enumerate the set of clauses to create the sequence of DRS symbols, where clauses are separated by a special token. The set of clauses is provided by the experimental dataset. Parsing texts to their DRSs in clause form is called DRS clause parsing. 

\paragraph{Sequential Graph Form.}
Aiming to transform DRSs in graph form to a sequence of symbols, we enumerate the predicate nodes and generate their semantic relations, along with the positions of their satellites. As shown in Figure \ref{seq}, each predicate node generates an item, and this item is linearly described by its associated satellites.
For example, in the item ``\texttt{climb\_up.v.01} \texttt{Agent} -1 \texttt{Time} +1 \texttt{Theme} +2'', it is indicated that the predicate \texttt{climb\_up.v.01} has three satellites. One is positioned at -1 distance, labeled as \texttt{Agent}, another is at +1 distance, labeled as \texttt{Time}, and the third is at +2 distance, labeled as \texttt{Theme}. We enumerate the items from top to bottom to obtain the sequence of DRS symbols, where items separated by a special token. The order of the predicates (items) is in line with the order of the corresponding words. Parsing texts to their DRSs in graph form is called DRS graph parsing.

\subsection{Models}
We adopt sequence-to-sequence models as our parsing models. The models are fine-tuned with LoRA \cite{hu2021lora}, an effective parameter-efficient fine-tuning approach. 
Instead of training the entire model, LoRA adopts a different approach by keeping the pre-trained model frozen and introducing smaller trainable low-rank matrices into each layer of the model. These low-rank matrices can approximate the trainable parameters alongside the existing weight matrices in the pre-trained models, enabling the models to handle various tasks without the need to modify all of the parameters.

Formally, the layer's input $x$ is passed through the frozen part $W_0$ ($W_0 \in \mathbb{R}^{d\times k}$) and the trainable part $\Delta W$ ($\Delta W \in \mathbb{R}^{d\times k}$). The layer's output is computed:
\begin{equation}
    h = W_0x + \Delta Wx,
\end{equation}
where the trainable part can be approximated by two low-rank matrics $B$ and $A$:
\begin{equation}
    \Delta W = BA,
\end{equation}
where $B \in \mathbb{R}^{d\times r}$, $A \in \mathbb{R}^{r\times k}$, and $r \ll \mathrm{min}(d,k)$. Following the previous work \cite{hu2021lora}, we apply the trainable approximation to the query and value parameters within the attention layers, rather than the entire models.

\subsection{Cross-Lingual Training}
The cross-lingual training is model-agnostic. Given the set of multilingual training data, $D = \{D_0, D_1, \ldots, D_n\}$, where $D_i, i \in[0,n]$ indicates the set of training instances in  language $i$, and $D_i = \{(X,Y)\}^m_i$ where $m$ is the size of the training data. The model parameters are initialized as $\theta_0$.
In each training step, a batch of instances is selected from $D$ without language identification. The batches are used to train the models, and the final models $\theta_n$ are obtained, where $n$ represents the step in the cross-lingual training process. While various selection strategies can be applied, in our experiments, we have employed a simple random selection approach.

Additionally, following the cross-lingual training, the models can be further fine-tuned on language-specific training data if available, with the aim of introducing language bias. For instance, to build a parser for language $i$, we fine-tune the model $\theta_n$ with the language-specific training data $D_i$, and obtain the final model $\theta_{n'}^i$ for language $i$.



\section{Experiments}
We conduct experiments to investigate the effectiveness of cross-lingual training and to show the performance of the final models enhanced by this training approach.

\subsection{Experimental Settings}
\paragraph{Benchmarks.} We conduct our experiments on two benchmarks: the Parallel Meaning Bank (PMB; \citealt{abzianidze-etal-2017-parallel}) in versions 3.0.0 and 4.0.0. These benchmarks include English, German, Italian, and Dutch data annotated with DRSs in both clause and graph forms. For each language, the data is automatically annotated and categorized into gold and silver subsets. The gold data is fully corrected by human while the silver data is partially corrected. Following the previous works \cite{van-noord-etal-2018-exploring}, we divided the gold data into train, development, and test data, while all the silver data is utilized for training purposes.\footnote{In PMB 3.0.0, only English and German data has gold training data. Only PMB 4.0.0 provides the DRS graph annotations.} A summary of the data is presented in Table \ref{data}.

\paragraph{Model Settings.} We base our models on mT0, a sequence-to-sequence pre-trained language model \cite{muennighoff-etal-2023-crosslingual}, with the model card \texttt{mT0-large}.\footnote{\url{https://huggingface.co/bigscience/mt0-large}} In our setup, LoRA matrices are exclusively added to the query and value parameters within all the attention layers, with a fixed rank ($r$) of 32. A summary of the models used in our experiments is provided below:
\begin{itemize}
    \item \textbf{Base}~~~~The models are trained exclusively on the train data. If there is no gold train data, the models are trained on silver data instead.
    \item \textbf{Base+}~~~~The models take a two-step training process. In the first step, they are trained on both the train data and the silver data. Then, the models are fine-tuned on the train data in the second step.
    \item \textbf{Cross-lingual}~~~~The models are trained using the cross-lingual method without language identification.
    \item \textbf{Cross-lingual+}~~~~\textbf{Cross-lingual} is further fine-tuned on the language-specific train data. If there is no train data, we use the silver data.
\end{itemize}

\begin{table}[!tp]
    \centering
    \begin{tabular}{clrrrr}
    \toprule
         & & \multicolumn{1}{c}{en} & \multicolumn{1}{c}{de} & \multicolumn{1}{c}{it}  & \multicolumn{1}{c}{nl}  \\
         \midrule
\multirow{4}{20pt}{PMB 3.0.0}  & silver & 97,598 & 5,250& 2,693& 1,222\\
& train & 6,620 & 1,159 & /& /\\
& dev & 885 & 417 & 515 & 529\\
& test & 898 & 403& 547& 483\\
\midrule
& all & 106,001 & 7,229 & 3,755 & 2,234\\
\midrule
\midrule
\multirow{4}{20pt}{PMB 4.0.0}  & silver & 127,303& 6,355 & 4,088& 1,440\\
 & train & 7,668 &1,738 &685 & 539\\
& dev & 1,169& 559&540 & 437\\
& test & 1,048&547 &461 & 491\\
\midrule
& all & 137,188 & 9,199 & 5,774 & 2,907\\
\bottomrule

    \end{tabular}
    \caption{Statistics of the data in experiments.}
    \label{data}
\end{table}

\begin{table*}[!tp]
    \centering
    \begin{tabular}{lcccccccc|cc}
    \toprule
     &\multicolumn{2}{c}{en} & \multicolumn{2}{c}{de} & \multicolumn{2}{c}{it} & \multicolumn{2}{c}{nl} & \multicolumn{2}{c}{average} \\
     \cmidrule{2-11}
     & F1  $\uparrow$ & IF $\downarrow$  & F1 $\uparrow$ & IF $\downarrow$& F1 $\uparrow$& IF $\downarrow$ & F1 $\uparrow$ & IF$\downarrow$ & F1 $\uparrow$ & IF$\downarrow$\\
     \midrule
     PMB 3.0.0 & \multicolumn{10}{c}{clause} \\
     \midrule
Base&87.75 & 2.49 & 78.66 & 2.88 & 76.94 & 0.39 & 67.91 & 3.40 & 77.82 & 2.29\\
Base+&86.26 & 1.47 & 80.11 & 1.44 & / & / & / & / & / & / \\
Cross-lingual&83.60 & \textbf{1.13} & 80.05 & \textbf{0.48} & \textbf{81.13} & 0.19 & \textbf{79.56} & \textbf{0.95} & 81.09 & \textbf{0.69}\\
Cross-lingual+~~~~&\textbf{88.60} & 1.47 & \textbf{83.23} & 1.20 & 80.07 & \textbf{0.00}& 78.98 & \textbf{0.95} &\textbf{82.72} & 0.90\\
     \midrule
     \midrule
     PMB 4.0.0 & \multicolumn{10}{c}{clause} \\
     \midrule
     Base & 88.67 & 3.42 & 81.75 & 3.40 & 79.25 & 0.93 & 70.05 & 5.26& 79.93 & 3.25\\
Base+ & \textbf{89.81} & 3.25 & 83.82 & 2.33 & 82.41 & 0.93 & 73.44 & 5.26 & 82.37 & 2.94\\
Cross-lingual&84.44 & 3.51 & 83.77 & \textbf{0.89} & 83.43 & \textbf{0.19} & 82.54 & \textbf{2.06} &83.55 & \textbf{1.66}\\
Cross-lingual+&89.65 & \textbf{3.17} & \textbf{85.66} & 1.97 & \textbf{85.04} & 0.37 & \textbf{83.86} & \textbf{2.06} & \textbf{86.05} & 1.89\\
     \midrule
     PMB 4.0.0 & \multicolumn{10}{c}{graph} \\
     \midrule
     Base & 95.16 & 1.28 & 90.09 & 3.76 & 89.52 & 2.59 & 85.81 & 5.95 & 90.14 & 3.40\\
Base+&95.81 & 1.03 & 92.11 & 1.25 & 91.16 & 1.11 & 88.47 & 2.97 & 91.89 & 1.59\\
Cross-lingual&93.81 & \textbf{0.17} & 92.31 & 0.18 & 92.37 & \textbf{0.00} & 91.88 & \textbf{0.23} & 92.59 & \textbf{0.14}\\
Cross-lingual+&\textbf{95.83} & 0.43 & \textbf{93.39} & \textbf{0.00 }& \textbf{93.00} & 0.74 & \textbf{92.62} & \textbf{0.23} & \textbf{93.71 }& 0.35\\
\bottomrule

    \end{tabular}
    \caption{
    Results on development data for DRS clause parsing and DRS graph parsing in both PMB 3.0.0 and PMB 4.0.0 benchmarks. The IF (\%) column indicates the percentage of ill-formed outputs. The best scores are highlighted in bold.}
    \label{dev_results}
\end{table*}
\paragraph{Training Settings.} 
All the models are trained with an initial learning rate of 0.001. The model Cross-lingual is trained in 20 epochs, while training of Base and Base+ and the fine-tuning part of Cross-lingual+ span 100 epochs. The optimizer used is AdamW \cite{loshchilov2018decoupled}, along with a linear learning rate scheduler. The batch size is 8.

\paragraph{Evaluation Metrics.} 
We employ F1 scores provided by \textsc{Counter} \cite{van-noord-etal-2018-evaluating} to evaluate the models for DRS clause parsing. For DRS graphs, which bears similarities to AMR graphs, we use F1 scores calculated by \textsc{SMATCH} \cite{cai-knight-2013-smatch} that is widely used for the evaluation on AMR parsing. Additionally, we present the percentage of ill-formed outputs (IF).

\subsection{Results of Cross-Lingual Training}
In this section, we compare the models according to the parsing accuracy measured by F1 scores.
\paragraph{Cross-lingual VS Base.}
As shown in Table \ref{dev_results}, on average, Cross-lingual outperforms Base in both DRS clause parsing and DRS graph parsing, consistently generating a higher quality of well-formed DRSs, particularly in German, Italian, and Dutch. In PMB 3.0.0, where gold training data is absent for Italian and Dutch, Cross-lingual achieves significant F1 improvements of 4.19\% and 11.65\%, respectively, in DRS clause parsing. Even in PMB 4.0.0, which provides limited-scale gold training data for these languages, Cross-lingual still demonstrates significant F1 improvements, with gains of 4.18\% and 12.49\% in DRS clause parsing and 2.85\% and 6.07\% in DRS graph parsing, for Italian and Dutch, respectively.

\paragraph{Cross-lingual VS Base+.}
In order to improve Base, Base+ is trained on a larger dataset comprised of silver data and gold train data and is then fine-tuned with the gold train data.
As shown in Table \ref{dev_results}, Cross-lingual outperforms Base+ on average, with notable F1 improvements of 1.02\% and 9.1\% in DRS clause parsing for Italian and Dutch, respectively. Additionally, Cross-lingual achieves 1.21\% and 3.41\% F1 improvements in DRS graph parsing for Italian and Dutch, respectively. However, Cross-lingual does not perform as well as Base+ in the case of English.

\paragraph{Resource-rich Languages.} 
As shown in the English part of Table \ref{dev_results}, Cross-lingual underperforms Base and Base+ in English. One reason is that Cross-lingual fine-tunes pre-trained language models (PLMs) using multilingual training data to reach a universal optimization point. However, this point is marginally lower than the language-specific optimization point achieved with language-specific training data, particularly in resource-rich languages like English.

\begin{table*}[!tp]
    \centering
    \begin{tabular}{lcccccccc}
    \toprule
     &\multicolumn{2}{c}{en} & \multicolumn{2}{c}{de} & \multicolumn{2}{c}{it} & \multicolumn{2}{c}{nl}  \\
     \cmidrule{2-9}
     & F1  $\uparrow$ & IF $\downarrow$  & F1 $\uparrow$ & IF $\downarrow$& F1 $\uparrow$& IF $\downarrow$ & F1 $\uparrow$ & IF$\downarrow$\\
     \midrule
     PMB 3.0.0 & \multicolumn{8}{c}{clause} \\
     \midrule
Neural-Boxer \cite{van-noord-etal-2018-exploring} & 88.9 & \textbf{0.2} & 81.9 & \textbf{0.2}  & \textbf{80.5} & \textbf{0.1} & 71.1 & 0.7 \\
BiLSTM-Char \cite{wang-etal-2021-input} & 88.1 & / & / & / & / & / & / & /  \\
\midrule
Cross-lingual+ & \textbf{89.1} & 2.0 & \textbf{82.7} & 1.2 & 80.2 & \textbf{0.1} & \textbf{80.1} & \textbf{0.2}\\
     \midrule
     \midrule
     PMB 4.0.0 & \multicolumn{8}{c}{graph} \\
     \midrule
     UD-Box \cite{poelman-etal-2022-transparent} & 81.8 & \textbf{0.0} & 77.5 & \textbf{0.0} & 79.1 & \textbf{0.0}  & 75.8 & \textbf{0.0} \\
Neural-Boxer \cite{poelman-etal-2022-transparent} &92.5 & 2.3 & 74.7 & 0.5 & 75.4 & 0.0 & 71.6 & 1.0\\
MLM \cite{wang-etal-2023-pre} & 94.7 & 0.3 & 92.0 & 0.4 & \textbf{93.1} & 0.2 & 92.6 & 0.6 \\
\midrule
Cross-lingual+ & \textbf{96.3} & 0.2 & \textbf{92.8} & 0.7 & 93.0 & 0.6 & \textbf{93.1} & 0.2\\
\bottomrule

    \end{tabular}
    \caption{
    Results on test data for DRS clause parsing and DRS graph parsing in both PMB 3.0.0 and PMB 4.0.0 benchmarks. The IF (\%) column indicates the percentage of ill-formed outputs. The best scores are highlighted in bold.
    }
    \label{test_results}
\end{table*}

\subsection{Results of Monolingual Fine-Tuning}
\paragraph{Base VS Base+.}
Compared to Base, Base+ is pre-trained on additional auto-generated training data. As shown in Table \ref{dev_results}, on average, Base+ consistently outperforms Base in both DRS clause and graph parsing in PMB 4.0.0. In German, Italian, and Dutch, Base+ achieves substantial F1 improvements in DRS clause parsing, with gains of 2.07\%, 3.16\%, and 3.39\%, respectively. Additionally, in DRS graph parsing, it records improvements of 2.02\%, 1.64\%, and 2.66\% in these languages. In English, Base+ attains marginal improvements, with gains of 0.54\% in DRS clause parsing and 0.65\% in DRS graph parsing.

\paragraph{Cross-lingual VS Cross-lingual+.}
Compared to Cross-lingual, Cross-lingual+ fine-tunes Cross-lingual with the corresponding monolingual data for language-specific DRS parsing. As shown in the PMB 4.0.0 part of Table \ref{dev_results}, Cross-lingual+ consistently outperforms Cross-lingual on average. In German, Italian, and Dutch, monolingual fine-tuning leads to marginal improvements in DRS clause parsing, with respective F1 score gains of 1.89\%, 1.61\%, and 1.32\%. Similarly, in DRS graph parsing, improvements of 1.08\%, 0.6\%, and 0.74\% are obtained for these languages.

\paragraph{Resource-rich Languages.}
As shown in the English part of Table \ref{dev_results}, when compared to Base, Base+ shows marginal improvements in DRS clause and graph parsing in PMB 4.0.0. When compared to Cross-lingual, Cross-lingual+ achieves significant improvements in English DRS clause and graph parsing, with respective F1 score gains of 5.21\% and 2.02\% in PMB 4.0.0. While Cross-lingual gets lower performance of English DRS parsing compared to Base and Base+, Cross-lingual+, with the addition of monolingual fine-tuning, pushes up the performance. This demonstrates the importance of monolingual data in building DRS parsers for resource-rich languages.

\subsection{Well-Formed Results}
In this section, we compare the models according to the well-formed outputs measured by IF scores. 

As shown in Table \ref{dev_results}, while some models achieve superior performance, resulting in higher F1 scores, they do not consistently produce well-formed DRSs. For example, Cross-lingual, despite generating DRSs of slightly lower quality compared to Cross-lingual+, produces fewer ill-formed DRSs on average. It appears that Cross-lingual, without monolingual fine-tuning, tends to generate well-formed DRSs, potentially at the cost of overall accuracy.

Despite this, Cross-lingual and Cross-lingual+ can achieve the better performance on DRS parsing by consistently generating higher-quality, well-formed DRSs when compared to Base and Base+. We believe that the proposed cross-lingual training method can confidently yield better results in DRS parsing.

\subsection{Comparisons with SOTA Systems}

We compare our final model, Cross-lingual+, with previous works in the standard benchmarks PMB 3.0.0 and PMB 4.0.0. The results are shown in Table \ref{test_results}. In DRS clause parsing, despite a slight decrease in performance due to a small percentage of ill-formed DRSs, our model achieves state-of-the-art results in English, German, and Dutch. Neural-Boxer \cite{van-noord-etal-2018-exploring,poelman-etal-2022-transparent} and BiLSTM-Char \cite{wang-etal-2021-input} preprocess both the input texts and the output DRS as sequences of characters, while our models operate at the word level.\footnote{Due to the lack of previous works on DRS clause parsing in PMB 4.0.0, our DRS clause comparison is exclusively based on PMB 3.0.0. Nonetheless, we provide the test results for DRS clause parsing in PMB 4.0.0 in Section 5.6.}

In DRS graph parsing, our model attains state-of-the-art results in English, German, and Dutch, while its performance on Italian data closely matches that of the SOTA model referred to as MLM \cite{wang-etal-2023-pre}, which builds a pre-trained language-meaning model for DRS graph parsing based on mBART \cite{tang-etal-2021-multilingual} by tuning all the parameters of the PLMs. Instead, our parsers only fine-tune the part of parameters. UD-Boxer \cite{poelman-etal-2022-transparent} adopts the Universal Dependency parsing to generate DRS graphs, ensuring the production of well-formed DRSs.

\begin{table}[!tp]
    \centering
    \begin{tabular}{lcccccccc}
    \toprule
     & EN & DE & IT & NL\\
     \midrule
     All & 89.61 & 85.08 & 85.94 & 85.38 \\
     \midrule
     DRS operator &  94.95 & 94.74 & 95.14 & 94.76\\
     Semantic Role & 88.56& 85.89& 89.67& 86.57\\
     Concept & 87.32 & 78.46 & 76.49& 78.46\\
     Synset-Noun & 90.64 & 85.44& 84.83& 86.43\\
     ~~~~-Verb & 77.14& 57.55& 55.62& 53.64\\
     ~~~~-Adjective & 80.63& 58.50& 50.39& 57.55\\
     ~~~~-Adverb & 87.40 & 74.42& 66.67& 80.60\\
\bottomrule
    \end{tabular}
    \caption{Fine-grained results (F1\%) on test data, given by Cross-lingual+ for DRS clause parsing in PMB 4.0.0.}
    \label{fine-grained}
\end{table}
\subsection{Detailed Analysis}
We conduct a variety of additional experiments to provide further insight into the final model.

\paragraph{Fine-grained Evaluation.}
\textsc{Counter} \cite{van-noord-etal-2018-evaluating} offers a comprehensive breakdown analysis for DRS clause parsing by assessing various aspects, including DRS operators, semantic roles, concepts, and synsets. Table \ref{fine-grained} shows the test results of our parser and demonstrates the strong ability of our parser to predict DRS operators, semantic roles, and senses of nouns. However, it faces challenges in distinguishing the senses of verbs, adjectives, and adverbs, especially in the case of German, Italian, and Dutch. While most predicates have only one common sense, verbs, adjectives, and adverbs exhibit a greater diversity of senses than nouns. This diversity in senses can pose challenges, confusing the parsers on the sense disambiguation. Therefore, it becomes essential to address the issue of sense disambiguation in order to construct accurate DRS parsers. 

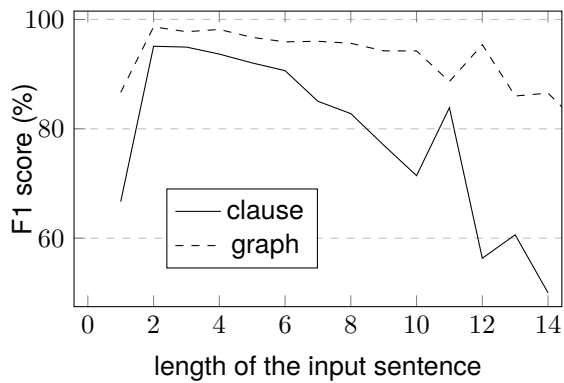
\begin{figure}[!tp]
\begin{tikzpicture}[scale=1]
\begin{axis} [
height = 5.5cm,
width = 8cm,
xlabel = length of the input sentence,
ylabel = F1 score (\%),
xmin=0,
xmax=14,
ymin=49,
ymax=100,
ytick pos=left,
legend style={at={(0.5,0.4)}},
ymajorgrids=true,
grid style=dashed,
y label style={at={(0.09,0.5)}},
enlargelimits=0.03
]

\addplot[black] coordinates{
( 1 , 66.67 )
( 2 , 95.1 )
( 3 , 94.95 )
( 4 , 93.67 )
( 5 , 92.05 )
( 6 , 90.64 )
( 7 , 85.03 )
( 8 , 82.78 )
( 9 , 77.02 )
( 10 , 71.43 )
( 11 , 83.87 )
( 12 , 56.330 )
( 13 , 60.61 )
( 14 , 50.0 )
};
\addlegendentry{clause}
\addplot[dashed] coordinates{
( 1 , 86.66666666666667 )
( 2 , 98.6522911051213 )
( 3 , 97.7620730270907 )
( 4 , 98.1763173375914 )
( 5 , 96.73348669076569 )
( 6 , 95.90498958895658 )
( 7 , 96.01589667163437 )
( 8 , 95.64541213063764 )
( 9 , 94.26157879405767 )
( 10 , 94.25010052271814 )
( 11 , 88.65546218487395 )
( 12 , 95.38461538461539 )
( 13 , 86.00000000000001 )
( 14 , 86.53295128939828 )
( 15 , 80.55555555555556 )
};
\addlegendentry{graph}
\end{axis}
\end{tikzpicture}
\caption{
The test results given by Cross-lingual+ for English DRS parsing in PMB 4.0.0 with input sentences of varying lengths.}
\label{lengths} 
\end{figure}

\paragraph{Evaluation on Text across Lengths.} 
Figure \ref{lengths} shows the F1 scores of English DRS clause and graph parsing by Cross-lingual+ on texts with varying lengths.\footnote{The trends observed in English serve as representative indicators of similar trends observed in other languages.} In both DRS clause and graph parsing, the F1 scores exhibit a consistent decrease as the input sentences become longer. This trend holds true for all input texts except for single-word inputs. The decline in F1 scores is primarily attributed to the challenge of over-generation in semantics. As an example, when the input text contains just one word, such as ``Hello'', indicating a simple spoken word, the parsers tend to generate additional predicates, like \texttt{female.n.02}, demonstrating a potential gender bias influenced by the training data. This over-generation phenomenon, particularly in shorter texts, leads to the lower F1 scores.

\paragraph{Errors in Ill-formed DRSs.}
Ill-formed DRSs significantly impede the performance of DRS parsers. In DRS clause parsing, errors within ill-formed DRSs can be categorized as follows: 1) illegal clause structure, such as an incorrect number of arguments for specific semantic relations; 2) generation of free variables, where some generated variables lack the interpretation with meanings. In DRS graph parsing, errors within ill-formed DRSs are summarized as follows: 1) inability to link distance symbols (e.g., +2) to specific items in the sequential graph. For instance, the total number of items in a sequential graph is 3, but parsers generate a distance symbol of +5, which falls outside the valid range; 2) over-generation of semantic relations, such as incorrectly generating a predicate with two Agents. We believe that avoiding ill-formed DRSs generation is necessary to improve DRS parsing.

\section{Conclusions}
We introduced a model-agnostic cross-lingual training approach, designed to leverage training data from various languages without specifying the language of the training instances. Our parsers are built on large-scale pre-trained language models with parameter-efficient fine-tuning. As a result, the parsers, enhanced by the proposed training methods, achieve state-of-the-art performance in the standard benchmarks for both DRS clause parsing and DRS graph parsing across multiple languages.
Furthermore, we conducted a variety of experiments to offer deep insights into the behavior of the parsers, aiming to inspire future research endeavors in the field of DRS parsing.

\section{acknowledgments}
This work was supported by Natural
Science Foundation of Yunnan Province of China (No.\ 202301CF070086) and the Open Project Program of Yunnan Key Laboratory of Intelligent Systems and Computing (No.\ ISC23Z01, 202205AG070003).
\section{Bibliographical References}\label{sec:reference}

\bibliographystyle{lrec-coling2024-natbib}
\bibliography{drs}

\appendix

\begin{table*}[!tp]
    \centering
    \begin{tabular}{c|cccccccc|cc}
    \toprule
     &\multicolumn{2}{c}{en} & \multicolumn{2}{c}{de} & \multicolumn{2}{c}{it} & \multicolumn{2}{c}{nl} & \multicolumn{2}{c}{average} \\
     \cmidrule{2-11}
     lr & F1  $\uparrow$ & IF $\downarrow$  & F1 $\uparrow$ & IF $\downarrow$& F1 $\uparrow$& IF $\downarrow$ & F1 $\uparrow$ & IF$\downarrow$ & F1 $\uparrow$ & IF$\downarrow$\\
     \midrule
     \midrule
     \multicolumn{11}{c}{clause in PMB 4.0.0} \\
     \midrule
 1e-3 & 86.63 & 2.67 & 85.41 & 1.46 & 86.09 & 0.22 & 84.29 & 1.43 & 86.36 & 1.44  \\
 1e-4 & \textbf{90.88} & \textbf{2.48} & \textbf{87.15} & \textbf{1.10} & \textbf{87.34} & \textbf{0.22} & 86.90 & 1.02 & \textbf{88.07} & 1.20 \\
 3e-5 & 90.69 & \textbf{2.48} & 86.59 & \textbf{1.10} & 87.15 & \textbf{0.22} & \textbf{86.93} & \textbf{0.81} & 87.84 & \textbf{1.15} \\
     \midrule
    \multicolumn{11}{c}{graph (sbn) in PMB 4.0.0} \\
     \midrule
1e-3 & 96.20 & 0.26 & 93.17 & 0.37 & 93.92 & 0.00 &	93.11 	&	0.41 & 94.10 &	0.26 \\

1e-4 & 97.00 &	\textbf{0.09} & 94.13 & \textbf{0.00} & \textbf{94.33} & \textbf{0.00} 	& 94.33 &	\textbf{0.00} & 94.95 & \textbf{0.02} 
\\

3e-5 &\textbf{97.03} & \textbf{0.09} & \textbf{94.26} & \textbf{0.00} & \textbf{94.33} & 0.00 &	\textbf{94.38} 	&	\textbf{0.00} & \textbf{95.00} & \textbf{0.02} \\
\midrule
    \multicolumn{11}{c}{graph (sbn) in PMB 5.1.0} \\
     \midrule
1e-3 &  92.22 	&	1.14 &	88.87 	&	0.18 	& 89.07 	&	0.65 	& 87.78 	&	1.22 	& 89.49 &	0.80 \\
1e-4 & \textbf{94.34} & \textbf{0.35} & \textbf{90.30} & \textbf{0.00} & 90.42 &	\textbf{0.00} & \textbf{89.99} & \textbf{0.00} & \textbf{91.26} & \textbf{0.09} \\
3e-5 & 94.30 & \textbf{0.35} & 90.25 & 0.18 &\textbf{90.64} & 0.43 & 89.56 & 0.41 & 91.19 & 0.34 \\
\bottomrule

    \end{tabular}
    \caption{
    Results on test data for DRS clause parsing and DRS graph parsing in PMB benchmarks. The IF (\%) column indicates the percentage of ill-formed outputs. The best scores are highlighted in bold.}
    \label{lr}
\end{table*}

\section{Additional Experiments}

We conduct additional experiments on different dataset. The base model is \texttt{mT0-large}. The models are pre-trained on the mixture of training data from all languages with 1e-3 learning rate in 30 epochs, and the batch size is 8. We choose the checkpoints of the last epoch for fine-tuning. 
Table \ref{lr} shows the Cross-lingual+ fine-tuned with different learning rates on language-specific training data in 100 epochs, and the batch size is 8. The models are trained on NVIDIA 3090 24G.

\end{document}